\documentclass[10pt,twocolumn,letterpaper]{article}

\usepackage{cvpr}
\usepackage{times}
\usepackage{epsfig}
\usepackage{graphicx}
\usepackage{amsmath}
\usepackage{amssymb}
\usepackage{subfigure}

 \cvprfinalcopy 

\def\cvprPaperID{2635} 

\ifcvprfinal\fi
\begin{document}

\title{SurfNet: Generating 3D shape surfaces using deep residual networks}


\author{
  Ayan Sinha \\
  MIT\\
 {\tt\small sinhayan@mit.edu} 
   \and
   Asim Unmesh\\
 IIT Kanpur\\
  {\tt\small a.unmesh@gmail.com} \\
   \and
  Qixing Huang\\
 UT Austin\\
 { \tt\small huangqx@cs.utexas.edu} \\
   \and
   Karthik Ramani \\
Purdue \\
{\tt\small ramani@purdue.edu} \\
}

\maketitle

\begin{abstract}
3D shape models are naturally parameterized using vertices and faces, \ie, composed of polygons forming a surface. However, current 3D learning paradigms for predictive and generative tasks using convolutional neural networks focus on a voxelized representation of the object. Lifting convolution operators from the traditional 2D to 3D results in high computational overhead with little additional benefit as most of the geometry information is contained on the surface boundary. Here we study the problem of directly generating the 3D shape surface of rigid and non-rigid shapes using deep convolutional neural networks. We develop a procedure to create consistent `geometry images' representing the shape surface of a category of 3D objects. We then use this consistent representation for category-specific shape surface generation from a parametric representation or an image by developing novel extensions of deep residual networks for the task of geometry image generation. Our experiments indicate that our network learns a meaningful representation of shape surfaces allowing it to interpolate between shape orientations and poses, invent new shape surfaces and reconstruct 3D shape surfaces from previously unseen images\footnote{Code available at https://github.com/sinhayan/surfnet}.

%
%

\end{abstract}

\section{Introduction}

The advent of virtual and augmented reality technologies along with the democratization of 3D printers has made it imperative to develop generative techniques for 3D content. Deep neural networks have shown promise for such generative modeling of 2D images \cite{DBLP:conf/icml/2015,DBLP:conf/nips/2014,DBLP:journals/corr/RadfordMC15}. Using similar techniques for creating high quality 3D content is at its infancy, especially because of the computational burden introduced by the $3^{rd}$ extra dimension \cite{ choy20163d, Girdhar2016,3dgan}.

Recent works in deep learning for 3D have argued for the redundancy of the $3^{rd}$ extra dimension as almost all of 3D shape information is contained on the surface. The authors of field probing neural networks \cite{li2016fpnn} address the sparse occupancy of voxel representations by developing adaptive 3D filters to reduce the cubic learning complexity. Following a similar argument, Sinha \etal propose to learn a 2D geometry image representation of 3D shape surfaces to mitigate the computational overhead of the $3^{rd}$ extra dimension \cite{Sinha2016}.
\begin{figure}[t]
\begin{center}
\includegraphics[width=0.98\linewidth]{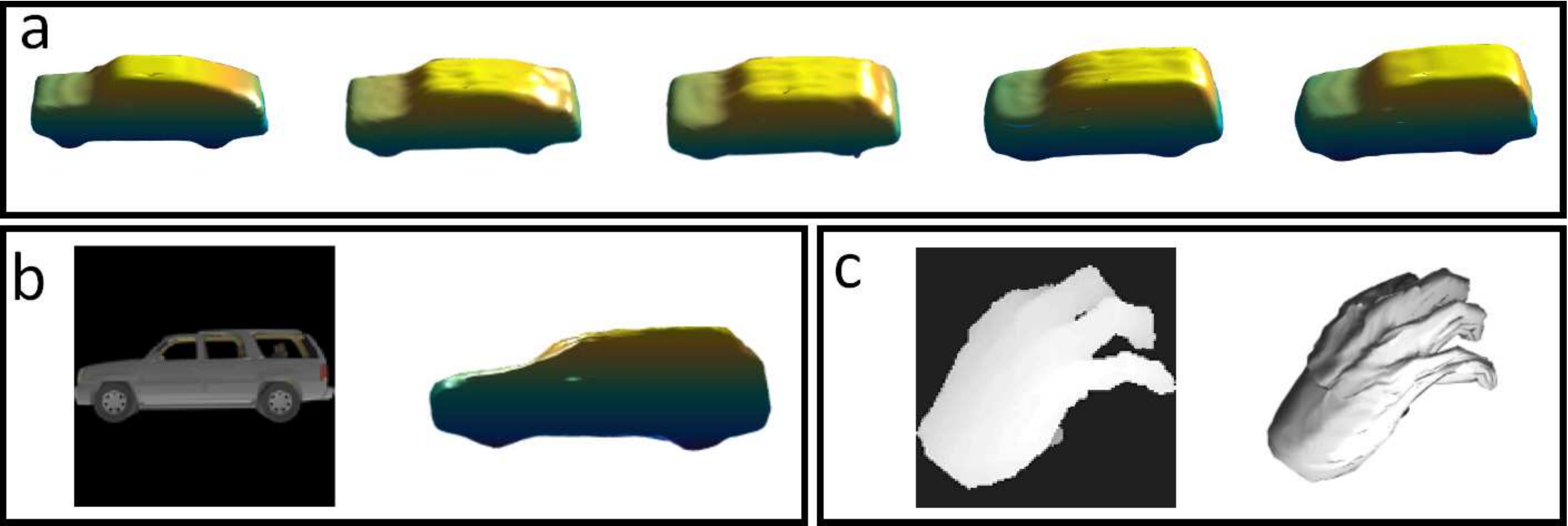}
\end{center}
   \caption{(a) 3D shape surface interpolation between original (left) and final (right) surface models with realistic intermediate styles is made possible by our generative deep neural network. (b) 3D rigid (or man-made) surface reconstruction from a RGB image, and (c) 3D non-rigid surface reconstruction from a depth image. The surfaces are constructed with implicit viewpoint estimation. }
\label{fig:firstfig}
\end{figure}

Here, we adopt the geometry image representation for generative modeling of 3D shape surfaces. Naively creating independent geometry images for a shape category and feeding them into deep neural networks fails to generate coherent 3D shape surfaces. Our primary contributions are: (1) A procedure to create consistent and robust geometry images for genus-0 surfaces across a shape category invariant to cuts and the intermediate spherical parametrization by solving a large scale correspondence problem, and (2) extending deep residual networks to automatically generate geometry images encoding the $x,y,z$ surface coordinates with implicit pose estimation and preservation of high frequency features for rigid as well as non-rigid shape categories. We demonstrate that neural networks trained using images or a parametric representation as inputs, and geometry images as outputs possess the ability to generate shape surfaces for unseen images, intermediate shape poses and interpolate between shape surfaces as shown in figure \ref{fig:firstfig}. Our paper is organized as follows. Section 2 discusses relevant work. Section 3 discusses the geometry image creation. Section 4 discusses the neural network architecture. Section 5 shows the result of our method, and section 6 discusses limitations and future work.

\section{Related Work}

Creating 3D content is an important problem in computer vision. Early works focussed on coherent synthesis of 3D primitives and surface patches \cite{Carlson:1982}. Recent approaches for assembly-based 3D shape creation from components use probabilistic models \cite{Chaudhuri:2011,Kalogerakis:2012}, or deep-learned models \cite{Huang:2015}. Estimates of wireframe for 3D objects are obtained by a 3D geometric object class model in \cite{zia2013detailed}. Kar \etal learn a deformable 3D model for shape reconstruction from a single image \cite{DBLP:conf/cvpr/KarTCM15}. Huang \etal show that joint analysis of image and shape collection enables 3D shape reconstruction from a single image \cite{Huang:2015}.

The success of deep learning architectures for generating images \cite{DBLP:conf/icml/2015, DBLP:conf/nips/2014} has resulted in extension of these techniques to generate models of 3D shapes. The authors of 3D ShapeNets \cite{wu20153d} perform pioneering work on using deep neural nets for 3D shape recognition and completion. Girdhar \etal \cite{Girdhar2016} learn a vector representation for 3D objects using images and CAD objects which are used for generating 3D shapes from an image. A volumetric denoising auto-encoder is demonstrated for 3D shape completion from noisy inputs in \cite{sharma16eccvw}. Choy \etal propose a 3D recurrent reconstruction neural network for 3D shape creation from single or multiple images \cite{choy20163d}. A probabilistic latent space of 3D shapes is learnt by extending generative-adversarial model of \cite{DBLP:conf/nips/2014} to the 3D domain in \cite{3dgan}. All these deep learning methods use 3D voxel representation for generating the 3D shapes. A conditional generative model is proposed in \cite{DBLP:journals/corr/RezendeEMBJH16} to infer 3D representation from 2D images. Although, this method can generate both 3D voxels or meshes, the mesh representation is limited to standard parameterizations which restrict shape variability. A 3D interpreter network is developed in \cite{3dinterpreter} which estimates the 3D skeleton of a shape.

Different from all above approaches, our thrust is to generate category-specific 3D point clouds representative of a surface instead of voxels to represent 3D objects. Our work is motivated by geometry image \cite{Gu:2002:GI:566570.566589} representation used for learning 3D shapes surfaces in \cite{Sinha2016}. Our neural network architecture is inspired by deep residual nets \cite{he2016deep} which have achieved impressive results on image recognition tasks, and by the architectural considerations in \cite{DB15} to generate chairs.

\section{Dataset Creation}

Our method to generate 3D shapes surfaces relies on a geometry image representation, \ie a remesh of an arbitrary surface onto a completely regular grid structure (see \cite{Gu:2002:GI:566570.566589} and supplement). Here, we consider car and airplanes to be prototypical examples of rigid, and the hand to be an example of a non-rigid shape. We detail the procedure to generate the geometry images and RGB or depth images required to train the 3D surface generating neural network.

\subsection{Non-rigid shapes} We use a kinematic hand model with 18 degrees of freedom (DOF), represented as $H(\theta)$, as standard
in hand pose estimation literature \cite{Sinha_2016_CVPR}. Here, $\theta$ denotes the set of 18 joint angle parameters. We generate synthetic depth maps by uniformly sampling each of the 18 joint parameters in the configuration space under dynamic and range constraints for joint angles. All hand mesh models contain 1065 vertices and 2126 faces wherein each vertex corresponds to the same point on the hand model, and the vertices have the same connectivity structure across all mesh models. The dataset covers a wide range of hand articulations from various viewpoints due to the 3 wrist rotation angles.

\begin{figure}[t]
\begin{center}
\fbox{\includegraphics[width=0.8\linewidth]{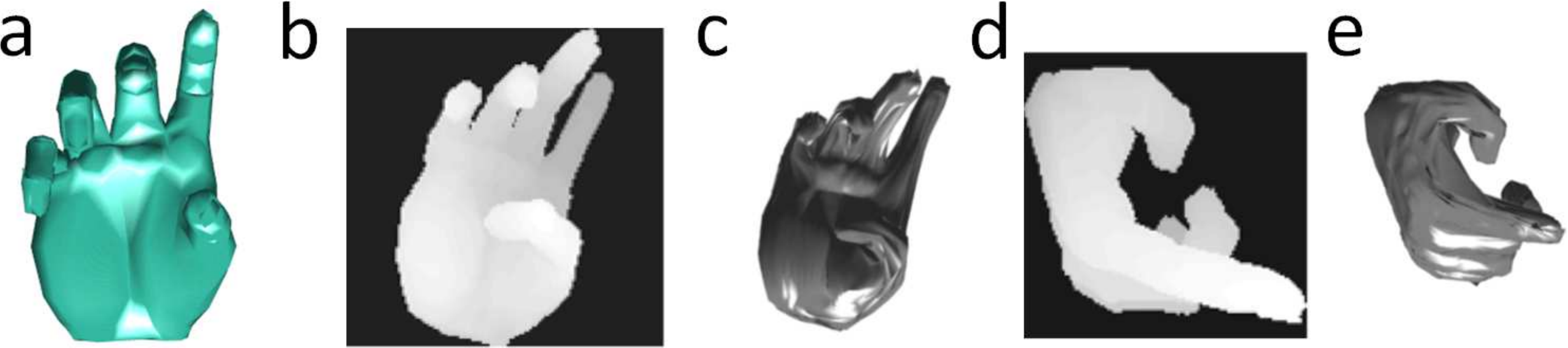}}
\end{center}
   \caption{(a) Hand mesh model. (b, d) Depth images rendered for different hand articulations. (c, e) Corresponding 3D surface plot of geometry images encoding the $x,y,z$ coordinates.}
\label{fig:lhand}
\end{figure}

\begin{figure}
\begin{center}
\fbox{\includegraphics[width=0.85\linewidth]{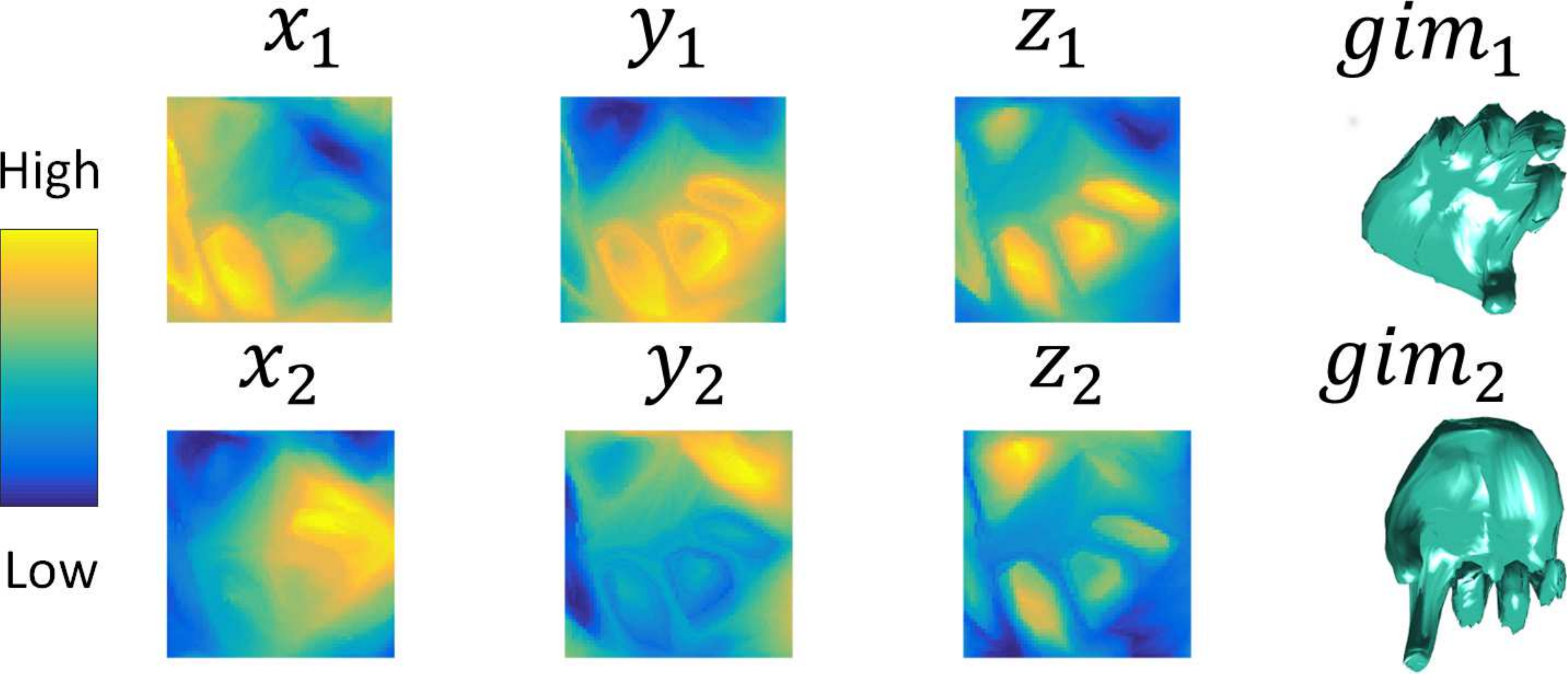}}
\end{center}
   \caption{Variation of geometry image due to hand articulation. The columns correspond to $x,y,z$-coordinate geometry image, and the encoded 3D plot by the geometry image, respectively. }
\label{fig:2hand}
\end{figure}

We generate 200,000 mesh files and store the 18 parameters, the 1065 vertex coordinates and the corresponding depth images. All depth images are normalized, cropped and resized such that the pixel with lowest depth has maximum intensity of 255, the hand is centered and the images are of size $128\times 128$.  Next, a randomly chosen mesh model is authalically and spherically parameterized using the method of \cite{Sinha2016}. Authalic spherical parametrization preserves the protruded features such as the fingers because the triangles on the original mesh model preserve their area on the derived parametrization. This spherical parametrization is converted to a flat and regular geometry image by first projecting onto an octahedron and then cutting it along 4 of its 8 edges (see \cite{praun2003spherical}). Geometry images can be encoded with any suitable feature of the surface mesh model such as curvature or shape signatures \cite{CGF:CGF12264} (also see supplement). As we are interested in reconstructing the 3D surface, all our geometry images are encoded with the $x,y,z$ values of points on the mesh model. These images are efficiently computed using the spherical parametrization of a single mesh model as all points across hand mesh models naturally correspond to each other. The geometry image of all meshes are of dimension $64\times64\times3$ corresponding to approximately $\approx 4000$ points area sampled on the hand. Methods for non-rigid shape correspondence such as \cite{DBLP:conf/iccv/ChenK15, Maron:2016:PRV:2897824.2925913,wei2016dense} can be used to develop dense one-to-one correspondence between mesh models when the correspondence information is unavailable. Figure \ref{fig:lhand} shows the mesh model, the rendered depth images and the 3D surface plots of $64\times 64\times 3$ geometry image. Figure \ref{fig:2hand} shows the variation of geometry images encoding $x,y,z$ coordinates for two different hand articulations. Observe that as the hand rotates, the intensities at the same spatial location in the geometry image of the $y$ coordinate, $y_1$ and $y_2$, are negatively correlated.

\subsection{Rigid or man-made shapes}
We create data for cars and aeroplanes mesh models from the ShapeNet database \cite{chang2015shapenet} to feed into our neural network architecture. We discuss the preprocessing steps and the correspondence development to create robust geometry image data for these synsets.

\noindent
\textbf{Preprocessing:} There are two constraints for the spherical parametrization technique of \cite{Sinha2016} to work on a mesh model. First, the surface mesh needs to follow the Euler characteristic. Almost all mesh models in ShapeNet do not follow the Euler characteristic, and hence, we first voxelize all mesh models at resolution $128\times128\times128$, and then create a $\alpha$-shape at $\alpha$-radius $\sqrt{3}$. This $\alpha$-radius preserves the holes and sharp edges in the derived surface mesh from the voxelized model. The surface mesh now follows the Euler characteristic after this preprocessing step. The second constraint for spherical parametrization is that the surface should be genus-0. We can use the heuristic proposed in \cite{Sinha2016} to generate non genus-0 surfaces by creating a topological mask in addition to the $x,y,z$ geometry images. However, for the sake of simplicity  we remove all mesh models derived from the $\alpha$-shape criterion with non-zero genus. We smooth the remaining mesh models using Laplacian smoothing to remove discretization errors.

\begin{figure}[t]
\begin{center}
\fbox{\includegraphics[width=0.8\linewidth]{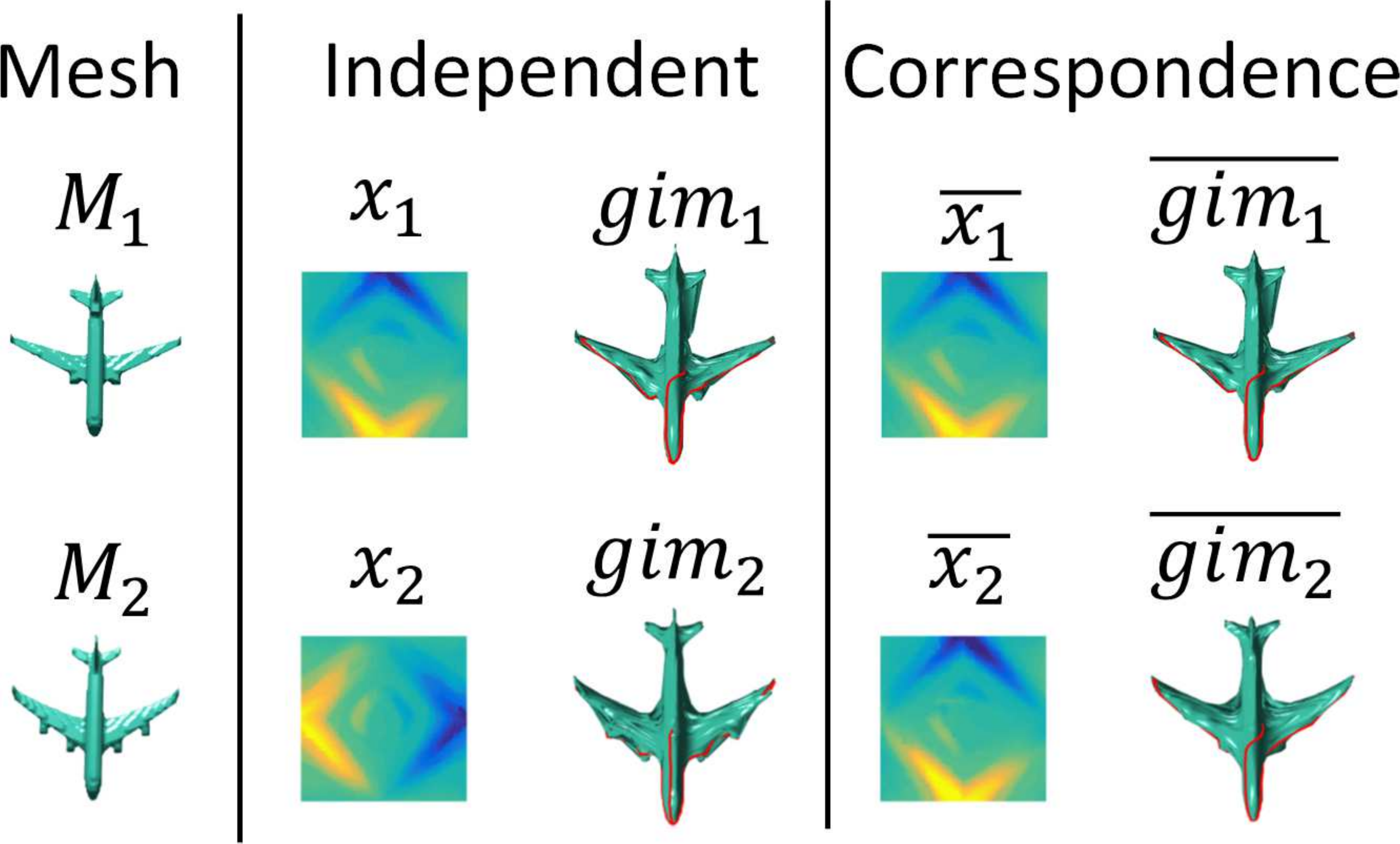}}
\end{center}
   \caption{Geometry images created by (1)independent parametrization of the two airplane models, and (2) By developing correspondence between the airplane meshes. }
\label{fig:gim1}
\end{figure}

\noindent
\textbf{Correspondence:} A naive strategy to create geometry image of $x,y,z$ coordinates on surface mesh models is to independently perform authalic spherical parametrization for all mesh models in a synset and then use these independent parameterizations to create a geometry image (details in supplement). However, such an approach suffers from severe limitations during learning with convolutional neural networks as: (1) The spherical parametrization is derived from area flow and cuts are defined \emph{a posteriori} to the parametrization. Different cuts will lead to different geometry images related by rotations and translations. This is displayed in figure \ref{fig:gim1} for two airplane models registered in the same pose. Independent parametrization results in the geometry images of the $x$ coordinate to be rotationally related. A generative neural network outputs a geometry image and gets confused when the cuts, and hence, the resulting geometry image for a shape in the same pose are different. This is similar to tasking a neural network to generate an image of an upright object by showing it several instances of the object in arbitrary poses without any prior informing it about the gravity direction. (2) Area preserving parametrization will result in a component of a shape to occupy varying number of pixels in a geometry image for different shapes in the same class, for \eg, an aircraft with large wings will have more pixels dedicated to the wing in a geometry image as compared to one with small wings. The neural network will have to explicitly learn attention to a component of a shape and normalize it in order to offset this bias. Our experiments with feeding independently parameterized shapes into a neural network for shape generation led to poor results. These problems with geometry images generated by independent parametrization of shapes in a class are resolved by performing parametrization for a single shape in a class, and establishing correspondence of all other shapes to this base shape. Figure \ref{fig:gim1} shows that the pixel intensities are correlated in the geometry images of the $x$ coordinate after establishing correspondence between two airplane models. The surface cuts highlighted in red follow the same contour unlike the independent case.

\begin{figure}
\begin{center}
\fbox{\includegraphics[width=0.85\linewidth]{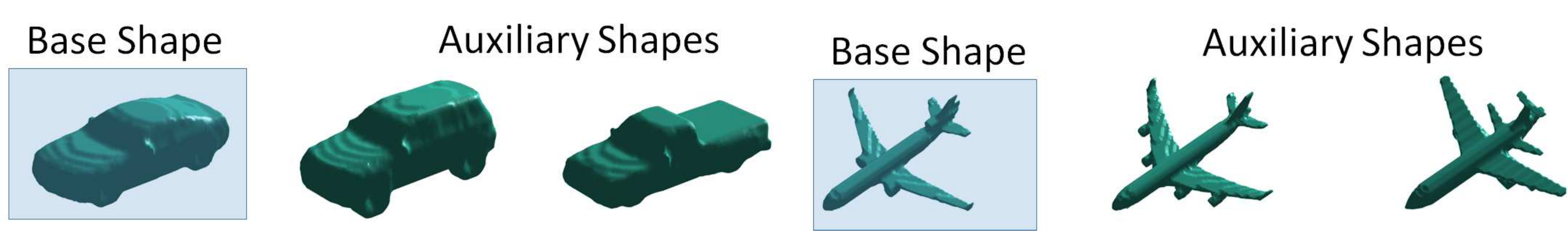}}
\end{center}
   \caption{Base and auxiliary shapes for the car and airplane models found by shape clustering. }
\label{fig:gim2}
\end{figure}

\begin{figure}[t]
\begin{center}
\fbox{\includegraphics[width=0.85\linewidth]{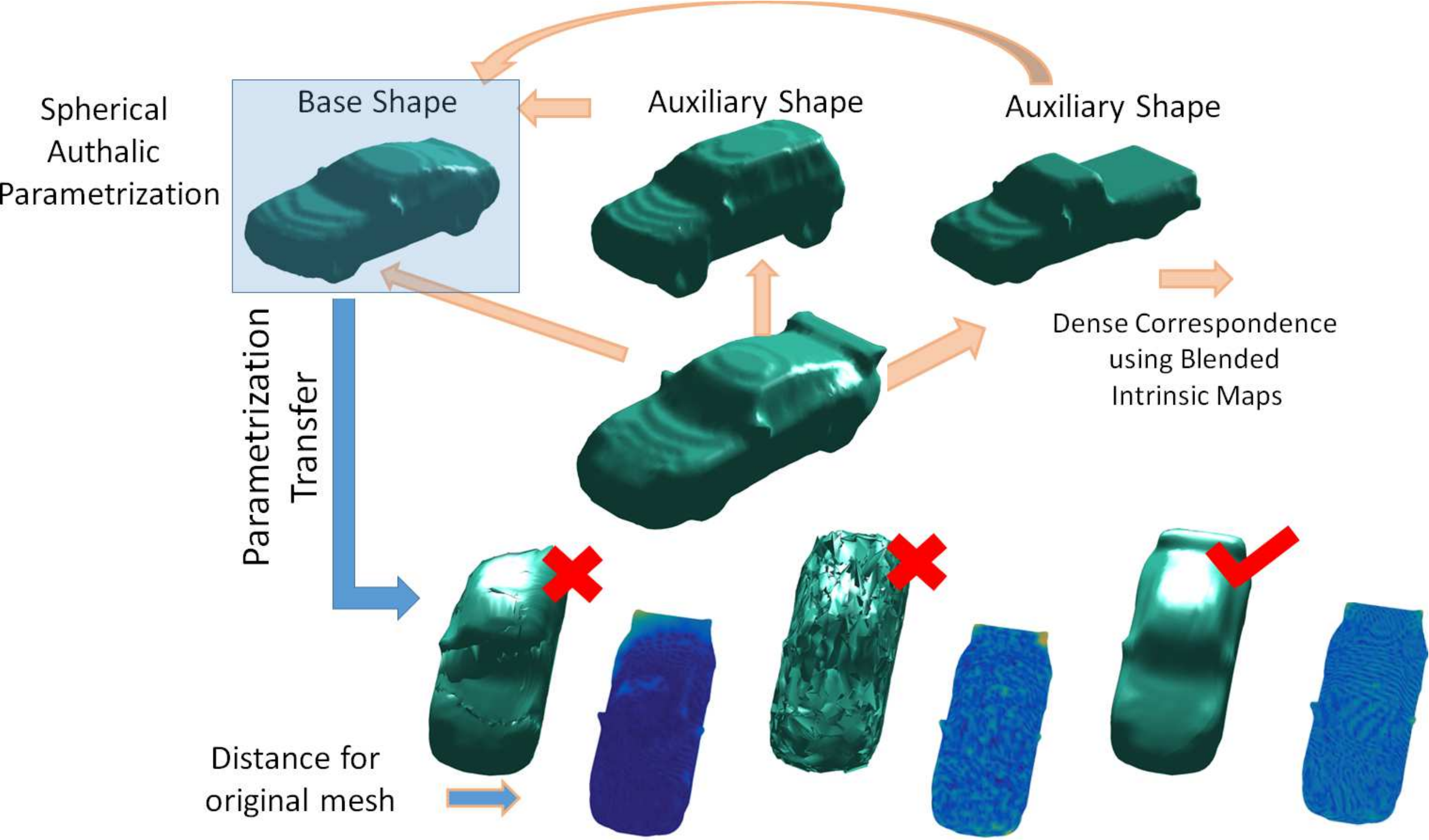}}
\end{center}
   \caption{We develop correspondences between mesh model and exemplar shapes to create consistent geometry images. The geometry image created using correspondences between central mesh model and axillary shape on top right shows best surface reconstruction lower than a threshold error, and subsequently used for training the neural network.}
\label{fig:corrcar}
\end{figure}

Establishing robust dense correspondence of surface meshes with high intra-class variability to a single mesh model is a hard problem. We resolve this problem by establishing dense correspondence of a shape to a few exemplar shapes in a class as follows. First, we create a shape similarity matrix using the distance between D2 descriptors \cite{Osada:2002:SD:571647.571648} of shapes in a class. Next, we perform spectral clustering \cite{Ng01onspectral} on the shape similarity matrix with $K=3$. The shapes closest to the cluster centroids are chosen as exemplars, and the cluster with maximum number of shapes is chosen as the base shape, $B$. The other two shapes serve as auxiliary shapes, $A$. Figure \ref{fig:gim2} shows the base and auxiliary shapes for the car and airplane synsets. We use blended intrinsic maps \cite{Kim:2011:BIM:1964921.1964974} to establish dense correspondence between a mesh model, $M$ and the three exemplar shapes. Dense correspondence between the base shape and the mesh model under consideration can be obtained directly $M\mapsto B$ or indirectly as $M\mapsto A \mapsto B$ by transferring correspondence information through an intermediate auxiliary shape using blended intrinsic maps as shown in figure \ref{fig:corrcar}. We perform spherical parametrization of the base mesh and use the correspondence information to create a geometry image of the mesh model, $M$ (see figure \ref{fig:corrcar}). We measure point-wise distances of surface points encoded in the geometry image to the original mesh model, and remove all models with average distance greater than a threshold. We are left with 691 car models and 1490 airplane models after dropping mesh models which have poor reconstruction using its geometry image evaluated by the average distance criterion.

%
%
%

\begin{figure}[t]
\begin{center}
\fbox{\includegraphics[width=0.98\linewidth]{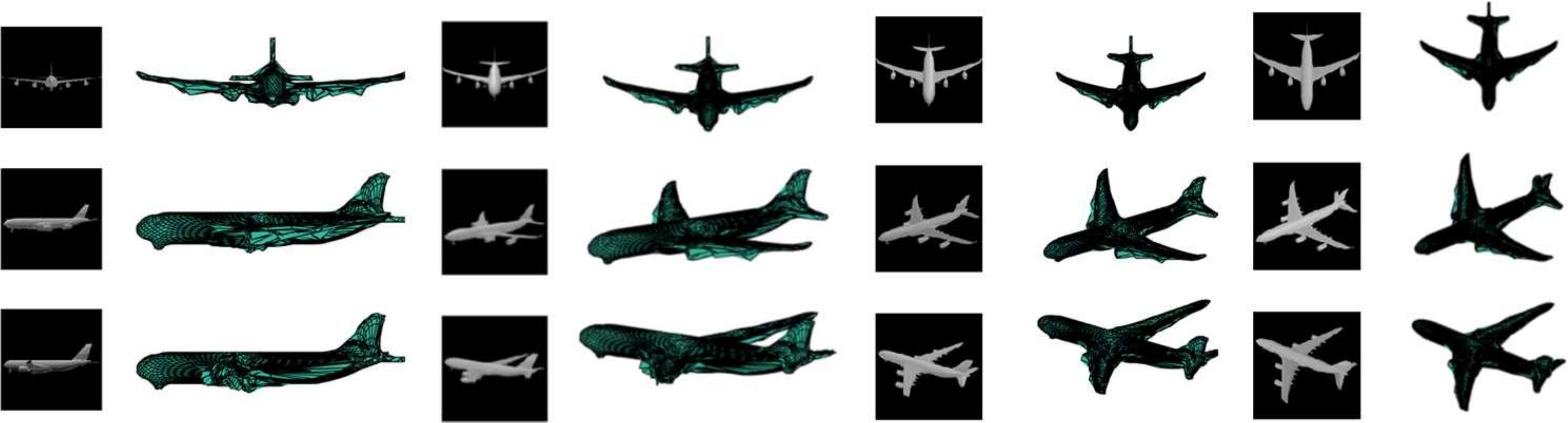}}
\end{center}
   \caption{Examples of rendered RGB images and corresponding surface plots encoded in the geometry images. }
\label{fig:gimimg}
\end{figure}

RGB images are rendered using Blender following the approach of \cite{Su_2015_ICCV} without background overlay. We consider 4 elevation angles $[0,15,30,45]$ and 24 azimuth angles in the range of 0 to 360 in intervals of 15. The corresponding geometry image is created by rotating the shape for the values of azimuth and elevation angles. Figure \ref{fig:gimimg} shows a few samples of the rendered images and corresponding geometry image. The images are of size $128\times128\times3$ and all geometry images are of size $64\times64\times3$ encoding $x,y,z$.

%

\begin{figure*}[t!]
  \centering
  \mbox{}
   \begin{subfigure}
  \centering
  \fbox{\includegraphics[width=0.48\linewidth]{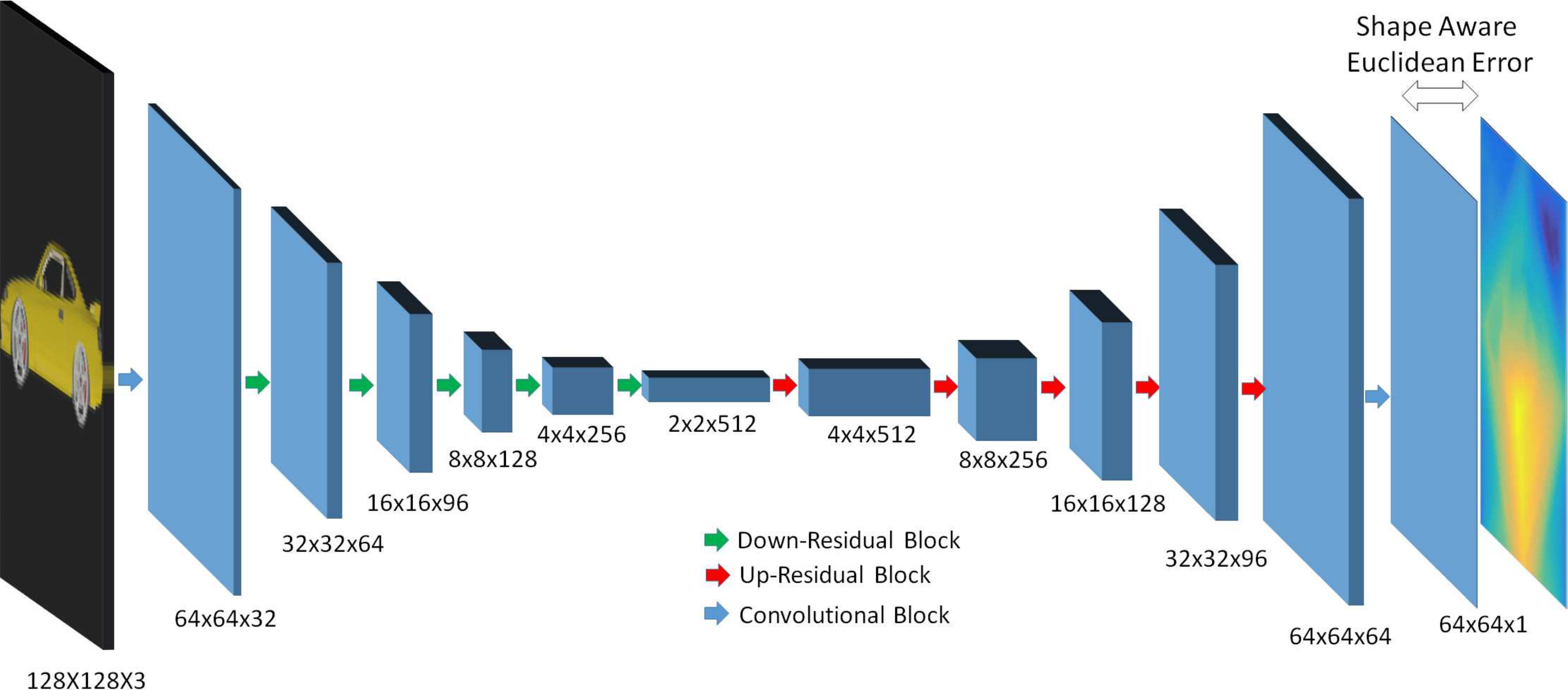} }
  \end{subfigure}
    \begin{subfigure}
  \centering
  \fbox{\includegraphics[width=0.47\linewidth]{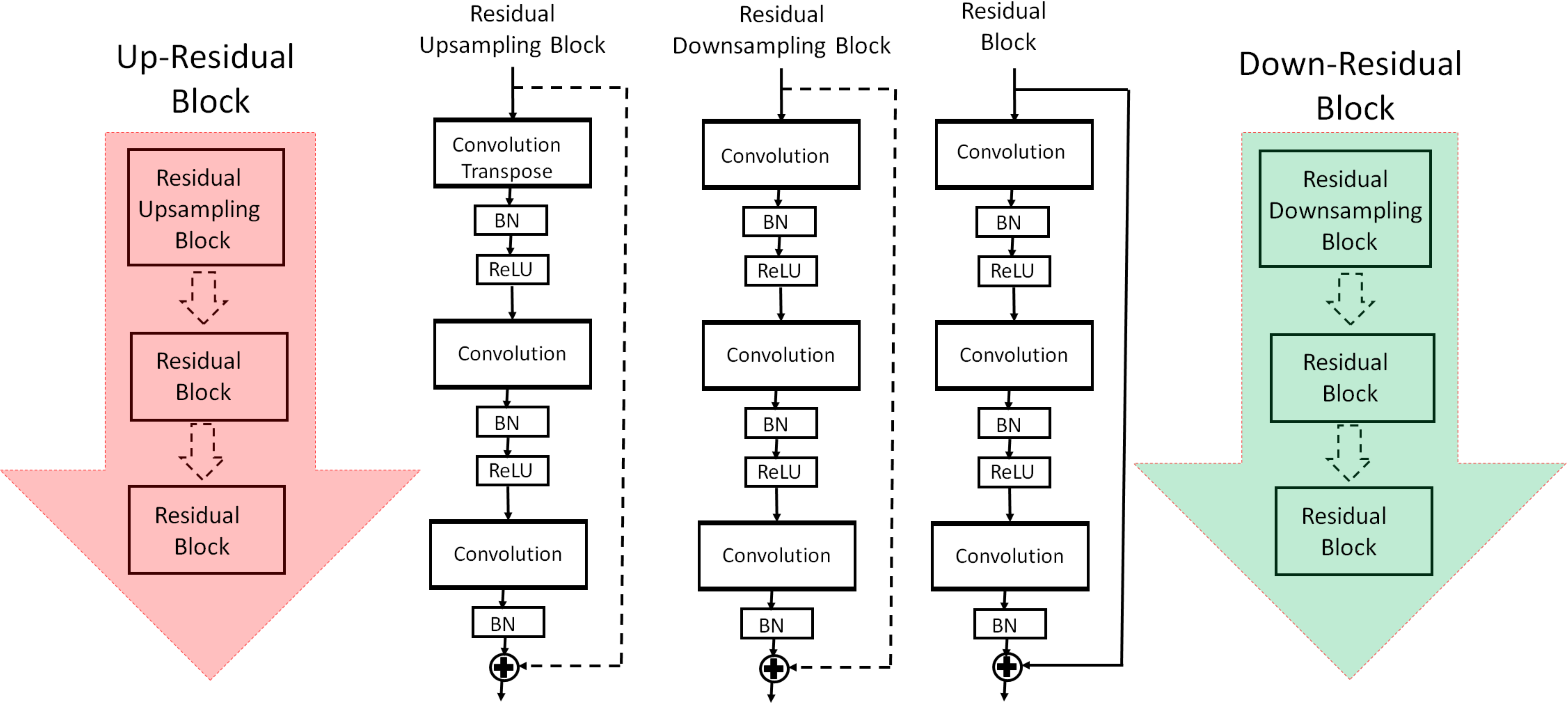}}
  \end{subfigure}

   \hfill \mbox{}
  \caption{Left: Architecture for generating a geometry image feature channel from an image. Right: The up (red) and down (green) residual building blocks of our network architecture composed of upsampling, downsampling and standard residual blocks shown in center. }
   \label{fig:gimarch}
\end{figure*}

\section{Deep Network Architecture}

Here we discuss the neural network architecture for rigid and non-rigid shapes for the following 2 scenarios: (1) Reconstructing 3D shape surface from a single image. (2) Generative modeling of 3D shape surface from a parametric representation.

\subsection{Reconstructing 3D surface from an image}

Inspired by the recent success of deep residual nets \cite{he2016deep} for image classification, we propose an extension of deep residual nets for image (here geometry image) generation. Figure \ref{fig:gimarch} left shows the network architecture for creating a feature channel of the geometry image. It is comprised of standard convolutions, the up-residual and down-residual blocks. The up and the down residual blocks increase and decrease the output size, respectively, and their composition is shown in figure  \ref{fig:gimarch} right. The up-residual block is composed of residual upsampling block followed two standard residual blocks, whereas the down-residual block is composed of residual downsampling block followed two standard residual blocks. The difference between the residual downsampling and upsampling blocks is that the first filter in the downsampling block is a convolution of size $3\times 3$,  1-padded with zeros and stride 2, whereas the first filter in the upsampling block is an convolution transpose (sometimes called deconvolution) of size $2\times 2$, 0-cropped and upsample 2. The solid side arrow in the standard residual block is a shortcut connection performing identity mapping. The dotted arrows in the up and down residual blocks are projection connections done using $2\times2$ convolution transpose and $1\times 1$ convolution filters, respectively. All convolution filters are of size $3\times 3$.  The input for the non-rigid database is a $128\times 128$ depth image, whereas the input for the rigid database is a $128\times 128\times 3$ RGB image. We tried directly generating all three $x,y,z$ feature channels of a geometry image using a single network and Euclidean loss between the network output and geometry image. However, the error of this network increases for a few epochs and then plateaus. This pattern persisted even after increasing the number of filters in the penultimate residual blocks before the output. Visually checking the output geometry images revealed that the network learnt a mean shape for a category. Instead, we learn each feature channel separately using three separate networks of the form shown in figure \ref{fig:gimarch} as each network devotes it entire learning capacity to learn either the $x,y$ or $z$ geometry image. The error of these networks generating a single feature channel decreased smoothly over epochs.

\begin{figure}[t]
\begin{center}
\fbox{\includegraphics[width=0.95\linewidth]{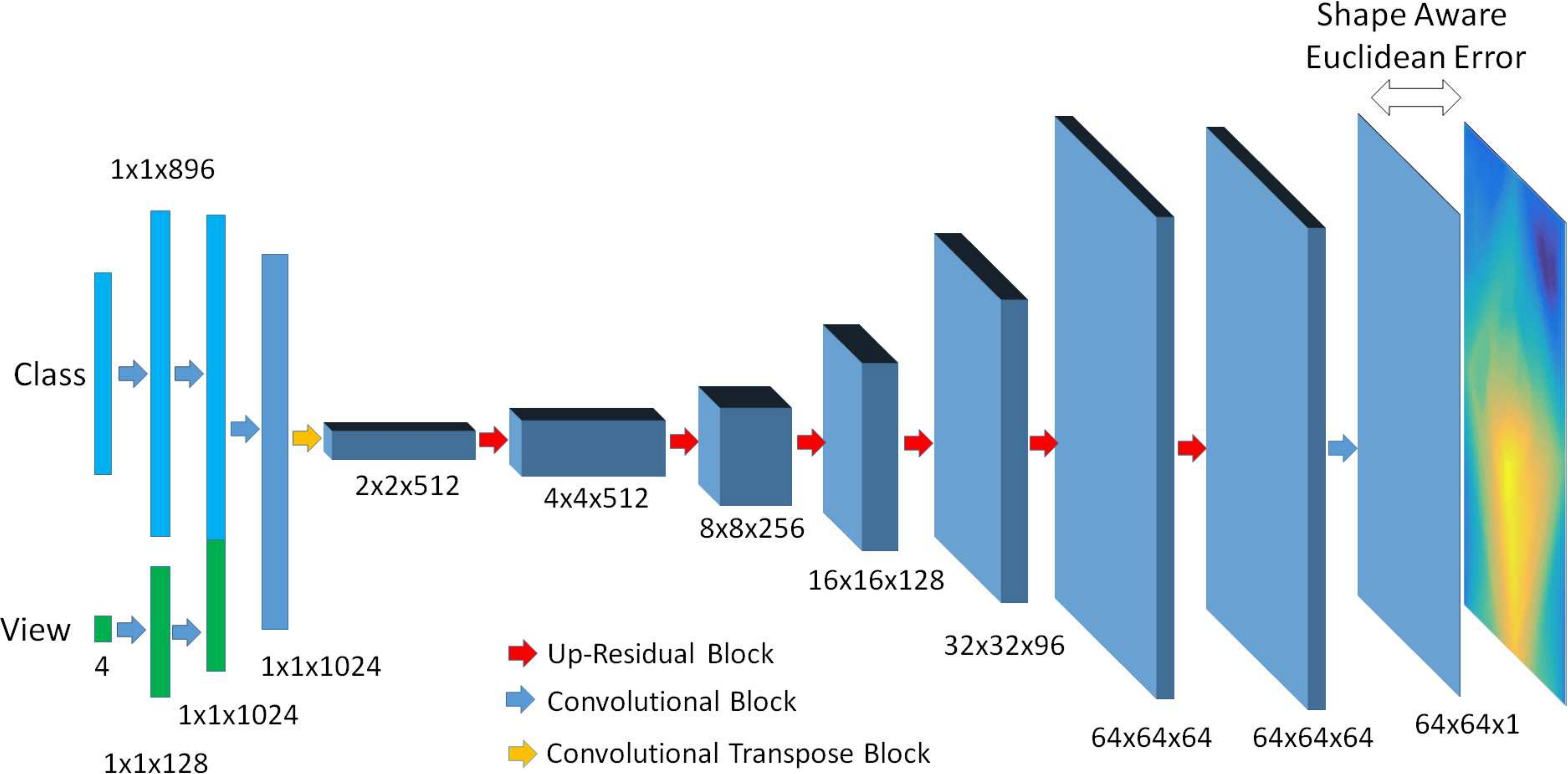}}
\end{center}
   \caption{Network architecture for generating a geometry image feature channel for rigid shapes from a one hot-encoded class label and view angle (in analogy to pose) parameters. }
   \label{fig:gimgen}
\end{figure}

Our next observation was that the network smoothed sharp edges especially in rigid datasets. So we employ a shape aware loss function of the form:
\begin{equation} \label{error}
min \sum_{(i,\theta)} \parallel |C^i|.(u_p(I^i(\theta))-g_p(\theta)) \parallel_2^2
\end{equation}
Here the minimization of weights in the neural network is over all training samples and configurations, $u_p$ is the output of the neural network learning feature $p$ with input $I^i(\theta)$, $i$ indicates the sample number, $\theta$ comprises the azimuth and elevation angles, $g^i_p(\theta)$ is the geometry image corresponding to feature $p$, sample $i$ and  angles $\theta$. $C^i$ is the geometry image of point-wise mean curvature for sample $i$. $C^i$ places higher weights on high curvature regions during learning and helps preserve sharp edges during reconstruction. We employ the same loss function for the non-rigid dataset.

\subsection{3D surface from a parametric representation}

\begin{figure}
\begin{center}
\fbox{\includegraphics[width=0.8\linewidth]{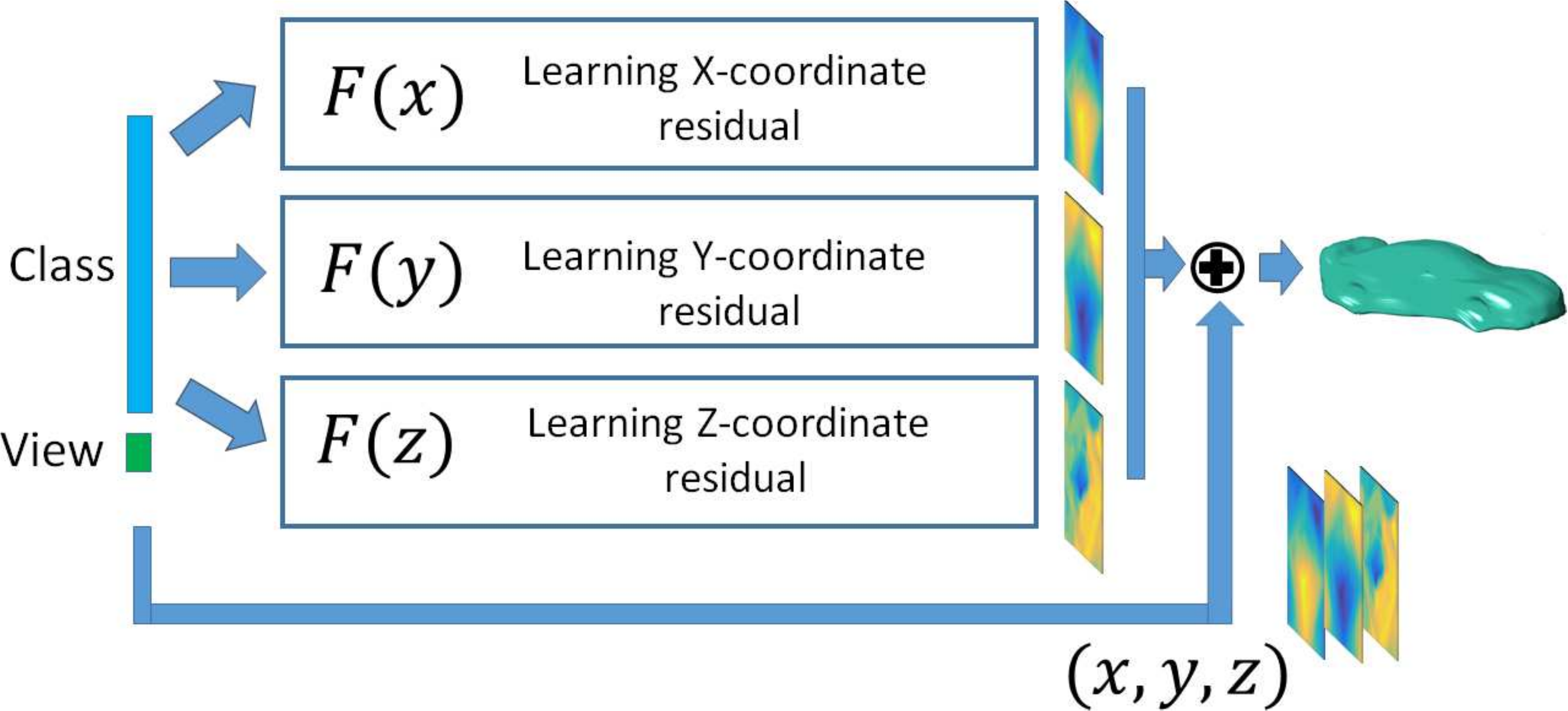}}
\end{center}
   \caption{Pipeline for generating a surface plot of a rigid shape from class and view parameters by summing the residual geometry image of $x,y,z$ coordinates to the base geometry image.}
\label{fig:rgen}
\end{figure}

We invert a residual network to generate 3D shape surfaces from a parametric representation. The parametric representation for the non-rigid hand is the 18 dimensional joint-angle vector, $H(\theta$). The parametric representation for the rigid datasets are two vectors: (1) $c$-a class label in one-hot encoding, (2) $\theta$- the azimuth and elevation view angles encoding shape orientation (each represented by their sine and cosine to enforce periodicity). Figure  \ref{fig:gimgen} shows the architecture for generating a geometry image for a single feature channel from a parametric representation for a rigid object. The architecture for the non-rigid hand is similar except without the view parameters and concatenation layer. The network comprises of up-residual blocks as described previously, and standard convolution and convolution transpose filters. The first two layers are fully connected. We again use separate networks to learn the $x,y,z$ geometry image.  We use the shape-aware loss function as described previously for independently generating the $x,y,z$-coordinate geometry image, and the hand surface is obtained by concatenating the three images into a single $64\times 64 \times 3$ geometry image. Figure \ref{fig:rgen} shows the pipeline for generating surface plots for the rigid datasets, and has a key difference to all other networks. As as we have explicit control over the $\theta$ parameters, we can generate a base shape with appropriate transformations due to $\theta$. In the spirit of residual networks, we generate a residual geometry image using the architecture shown in figure \ref{fig:gimgen}, and the final shape is derived by summing the residual geometry image of the $x,y,z$ coordinates to the geometry image of the base shape. We observed that learning residual geometry images led to faster convergence and better preservation of the high frequency features of the shape. We cannot perform residual learning on the hand as the global rotations due to the wrist angles are continuous, and not discretized in azimuth and elevation. We cannot perform residual learning on rigid shapes generated from an image as the $\theta$ parameters are implicit in RGB image.


\subsection{Training details}

We train our networks using MatConvNet  and a Nvidia GTX 1080 GPU. The learning rate was 0.01 and we decreased the learning rate by a factor of 10 after every 5 epochs. We trained the 3D reconstruction neural networks from a single image with 102 layers for 20 epochs, and the generative networks from a parametric representation with 65 layers for 15 epochs. The momentum was fixed at 0.9. All rectified linear units (ReLU) in up-residual blocks had leak 0.2. We experimented with geometry images of resolution $128\times 128$ (instead of $64\times 64$) and found no difficulties in learning, albeit at a larger training time. We used 80\% of the 200,000 hand models, 691 car models and 1490 airplane models for training and the rest were used for testing reconstruction from a single image. We manually pruned the rigid models to remove near-duplicate shapes and were left with 484 car models and 737 airplane models, all of which were used for training the generative models from a one-hot encoded vector.

\section{Experiments}

In this section, we first discuss generating 3D shape surfaces for the non-rigid hand model and then perform experiments on generating 3D shape surfaces for the rigid aeroplane and car datasets. We generate surfaces using a parametric representation and from an image.


\begin{figure}[t]
\begin{center}
\fbox{\includegraphics[width=0.8\linewidth]{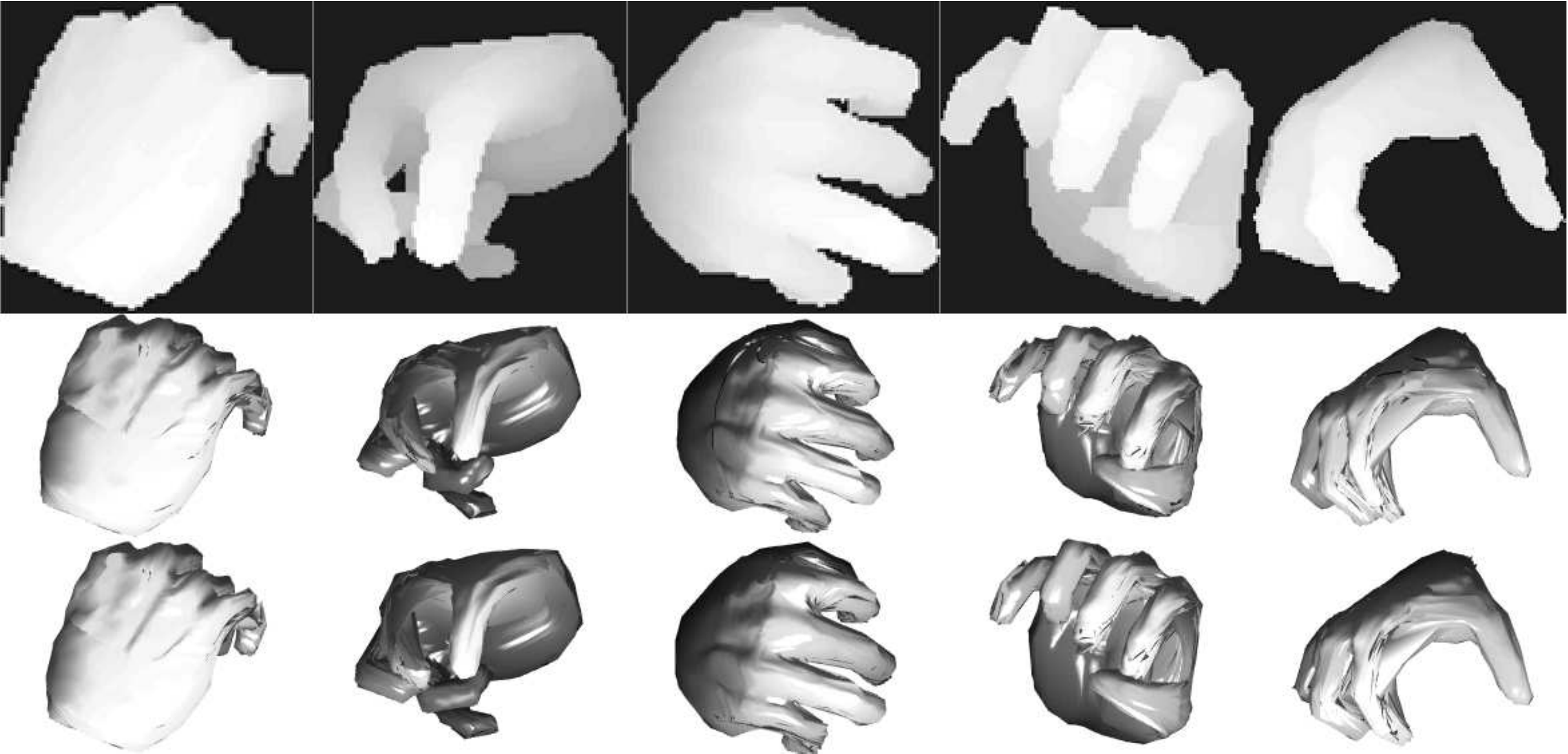}}
\end{center}
   \caption{Results on test dataset for reconstructing the 3D shape surface of the hand from a single depth image. The first row is the depth image, the second row is the ground truth and the third row is our reconstruction. }
\label{fig:nrgenhand}
\end{figure}

\begin{figure}
\begin{center}
\fbox{\includegraphics[width=0.78\linewidth]{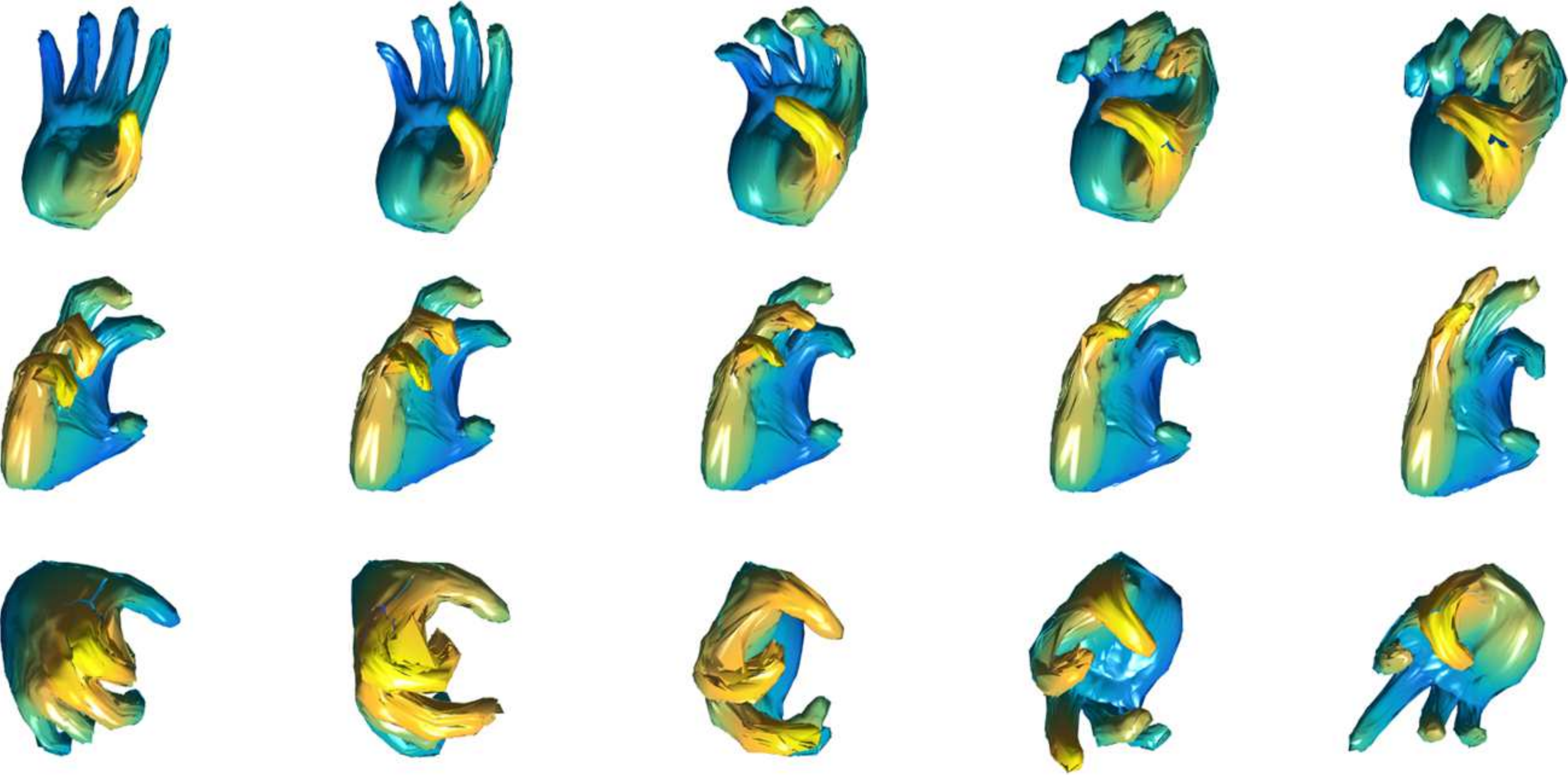}}
\end{center}
   \caption{Each row shows the 3D surface plots of geometry images created by our neural network by inputting uniformly spaced parametric joint angle vectors.}
\label{fig:nrgencontinuous}
\end{figure}

\begin{figure*} [t!]
\begin{center}
\fbox{\includegraphics[width=0.98\linewidth]{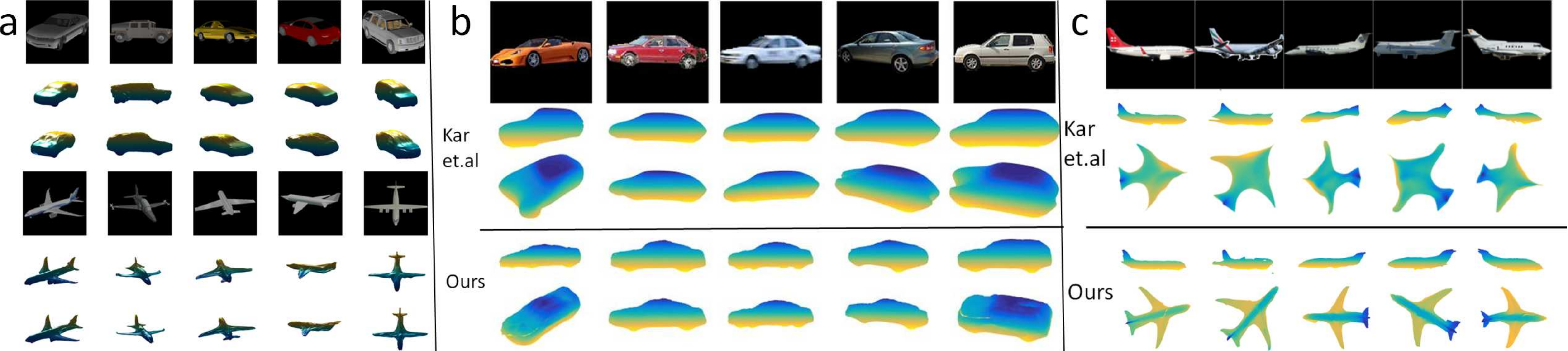}}
\end{center}
   \caption{3D reconstruction of rigid surfaces from a single RGB image. (a) Results on test dataset for reconstructing the 3D shape surface of cars (top) and airplanes (bottom) from a single RGB image. The first row is the depth image, the second row is the ground truth and the third row is our reconstruction for both categories. (b,c) Comparing our method to \cite{DBLP:conf/cvpr/KarTCM15} on the PASCAL 3D+ car (b) and aeroplane (c) dataset. We show the regressed viewpoint and an alternate viewpoint, for each 3D reconstruction to better reveal the quality of the methods. }
\label{fig:rgencar}
\end{figure*}

\subsection{Non-rigid shapes}

Figure \ref{fig:nrgenhand} shows few 3D surface plots of the generated geometry image by our neural networks on the test depth images. We see that it is able to recover the full articulation of the hand very close to the ground truth even in the presence of occlusion. For example, the middle finger is well approximated from the depth image in the second test case although it suffers from high occlusion. We also note that although we trained separate neural networks for generating the $x,y,z$ geometry image, the combined result shows good fidelity in terms of spatial localization. The supplement discusses quantitative evaluation on test datasets. These results are encouraging for hand tracking applications using a depth camera. Unlike standard methods which estimate the joint angle \cite{choi2015collaborative,Sinha_2016_CVPR} or joint position parameters \cite{sharp2015accurate,sun2015cascaded,tompson2014real}, we reconstruct the full 3D surface. Our approach has the potential to go beyond pose estimation and even map individual hand textures (using a texture geometry image) to provide an immersive experience in virtual and augmented reality applications, which we wish to explore in future work.

Next, we perform experiments on generative modeling of non-rigid shape surfaces from a parametric representation. We consider two cases. First, we create two random 15-dimensional vectors for the local joint angles, and fix the 3 global wrist angles. We then linearly interpolate each dimension of the 15-dimensional vector from the first to the second random value, and sample values at equal intervals. The first two rows of figure \ref{fig:nrgencontinuous} show the output 3D surface plots by inputting these interpolated joint angle vectors. We see that there is a smooth transition from the first pose to the second pose indicating that the neural network did not merely memorize the parametric representation, but instead discovered a meaningful abstraction of the hand surface. Second, we create two random 18 dimensional vectors, and uniformly sampled from the linearly interpolated joint-angle values from the first to the second vector. The third row of figure \ref{fig:nrgencontinuous} shows the output 3D surface plots for this setting. Again, we observe the same phenomenon of natural transition from the first to the second pose.

%

%

\subsection{Rigid or man-made shapes}
We first discuss 3D reconstruction of a rigid object, here cars and planes, from a single image and then from a parametric representation.

\noindent
\textbf{3D surface reconstruction from a single image:} Figure \ref{fig:rgencar}(a) shows 3D surface plots of the generated geometry image by our neural networks on the test RGB images of the car and aeroplane respectively. We see that our neural network is able to correctly  estimate both the viewpoint as well as the 3D shape surface from the RGB image for diverse types of cars and aeroplanes. Current deep learning methods are able to estimate either the viewpoint  \cite{Su_2015_ICCV}, or reconstruct pose-oblivious 3D objects \cite{choy20163d, Girdhar2016,3dgan} from an image, but not both. With the ability to directly regress the surface to the appropriate pose, our work serves as a promising step towards fully automatic 3D scene reconstruction and completion. We observe in figure \ref{fig:rgencar}(a) that the reconstructed surface preserves sharp object edges, however has trouble enforcing smoothness on flat regions such as the windshield of the cars. We hypothesize that this is due to independent generation of feature channels and can be removed by simple post-processing. We also observed that the neural network had difficulty reconstructing cars with low intensity features such as black stripes as it was unintelligible from the background. We see that the tails of aeroplanes in figure \ref{fig:rgencar}(a) are faithfully reconstructed even though the tails in the ground truth are noisy or incomplete due to poor correspondence. This is because the neural network learns a meaningful representation of the 3D shape category. The supplementary material provides additional quantitative and qualitative results on the test dataset. We also ran our learnt networks on the airplane and car categories of the PASCAL 3D+ \cite{6836101} dataset and qualitatively compare it with the method of \cite{DBLP:conf/cvpr/KarTCM15}. We cropped and resized the images using the ground truth segmentation mask and fed them into our networks. In addition to the segmentation mask, we allowed the Kar \etal method to have keypoint labels. Note that our method only outputs the point coordinates of the surface and not the full mesh. Figure \ref{fig:rgencar}(b,c) shows that our method is able to reconstruct the car and airplane surfaces with good accuracy with small artifacts near the geometry image boundary, whereas the Kar \etal method has trouble discriminating between hatchbacks and sedans, and the spatial extent of wings even with keypoint labels. However, our network failed to output coherent 3D reconstruction results on some images. These were mostly images with low contrast, poor texture or views beyond our training ranges of azimuth and elevation angles.


%

%

%
%
%
%

\begin{figure*} [t]
\begin{center}
\fbox{\includegraphics[width=0.95\linewidth]{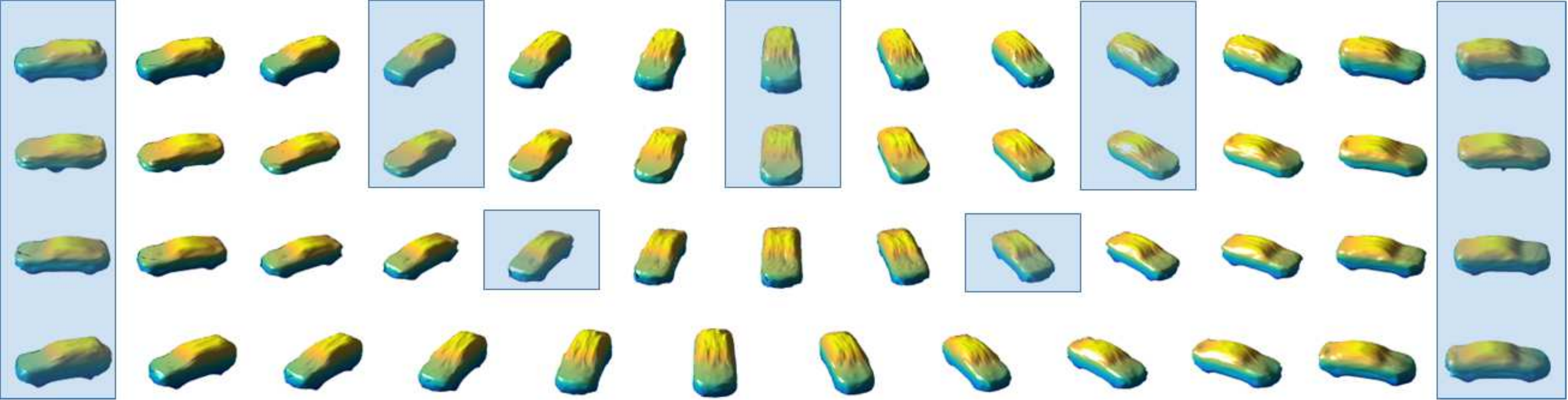}}
\end{center}
   \caption{Examples of interpolation of shape surfaces between azimuth angles. The highlighted shapes are in the training set. }
\label{fig:posevary}
\end{figure*}

\begin{figure*} [t]
\begin{center}
\fbox{\includegraphics[width=0.95\linewidth]{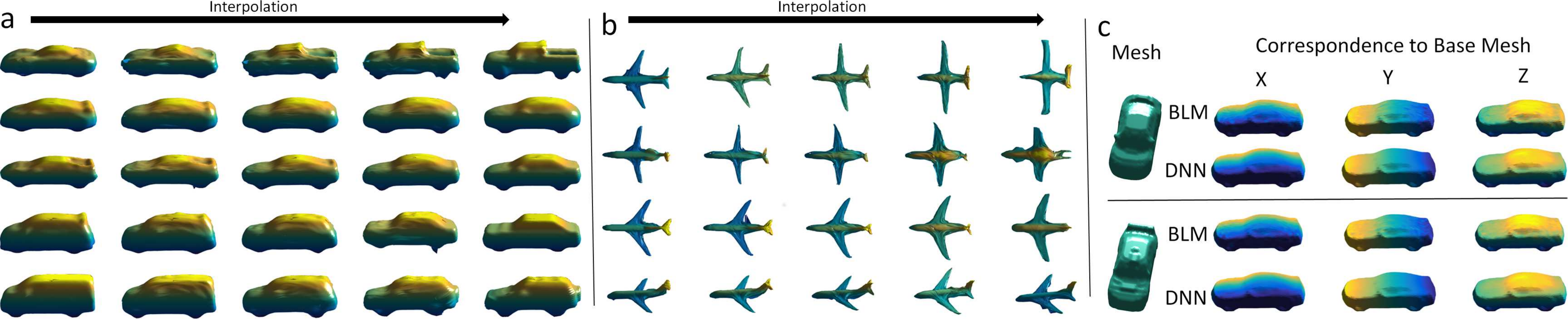}}
\end{center}
   \caption{(a,b) Shape surface interpolation, one morphing for each row, between the original model (left) and final model (right). (c) Rectifying correspondence information using our deep neural network architecture for 3D surface reconstruction from an image.}
\label{fig:masterres}
\end{figure*}

\noindent
\textbf{3D surface generation from a one hot encoding:} We discuss the advantage of creating residual geometry image approach instead of a direct image in terms of reconstruction error in the supplement. We first keep a constant one hot encoding and vary the training set size in terms of the azimuth angle. Figure \ref{fig:posevary} shows the results of interpolating 3d shape surfaces between azimuth angles wherein for each row, the highlighted shapes are in the training set and the remaining unseen 3D shape poses are generated by the deep residual network. This capability of the network to generate unseen intermediate poses reflects that the network internally learns a representation of 3D shape surfaces. This is further validated by linearly varying the one-hot encoded vector from one shape surface to another shape surface in addition to the azimuth angle, and the result is shown in the last row of figure \ref{fig:posevary}. We see that the network generates realistic intermediate surfaces in addition to varying the azimuth angle.

We further experiment the phenomenon of 3D surface interpolation between two shape surfaces in figure \ref{fig:masterres}(a,b). Each row shows morphing between two shape surfaces wherein the first and the last shapes are 3D surface reconstructions by the neural network for two different one-hot encoded vectors and the intermediate 3 shape surfaces are generated by inputting values [0.75,0.25], [0.5,0.5], [0.25,0.75] corresponding to the active codes in the vector. In figure \ref{fig:masterres}(a) we see that the shape surface varies smoothly between a convertible and a pickup truck ($1^{st}$ row), a sports car and a SUV ($2^{nd}$ row), and a van and a jeep ($5^{th}$ row) all while generating realistic intermediate car body styles. We observe the similar results when we linearly interpolate between one-hot encoded vectors for two airplane surfaces. In the first row of figure \ref{fig:masterres}(b) we see that the neural networks learns a consistent internal representation of airplane wings, and in the last row the same can be said about the tail. We expect 3D modelers to benefit from such a generative model to create new content.

%

\subsection{Correspondence}

Developing robust correspondences between a mesh model and a base mesh is an important step in our pipeline, but is fraught with challenges especially when the shape category has high intra-class variation. We demonstrate that the internal representation learnt by a deep neural network can help remove noise from correspondence information between two surface meshes as follows: (1) Pick a model with noisy or incorrect correspondence, (2) Render its image from an appropriate angle and feed it into the neural network for reconstructing shape surface from an RGB image. The output geometry image of point coordinates has one-to-one correspondence to the geometry image of the base mesh, which in turn establishes direct correspondences between the mesh models.  This is shown for two models from the car training set in \ref{fig:masterres}(c). Observe that the point-to-point correspondence (displayed separately in color for each coordinate) are noisy and non-smooth on the surface of the base mesh as determined by blended intrinsic maps (BLM). This noise reduces and the color gradient indicating fidelity of correspondence smoothes when we use the output of the deep neural network (DNN) to establish correspondence. This correction mechanism hints that we can use feedback from the neural network to rectify noisy correspondences in the training set and also incorporate additional models for training a neural network, similar in spirit to \cite{Oberweger:2015:TFL:2919332.2919832}.

\section{Limitations and Future Work}

We have proposed what may be the first approach to generate 3D shape surfaces using deep neural networks. One limitation of our current approach is that it is limited to genus-0 surfaces which we wish to remove in future work. We also wish to explore the proposed feedback mechanism to improve correspondences, or to use more sophisticated correspondence methods such as \cite{Huang:2013:CSM:2600289.2600314} to improve and increase the training set size. Developing neural networks capable of learning multiple shape categories and all feature channels simultaneously without degradation in performance is a promising research direction. We are encouraged by the generality of our approach to generate 3D rigid or man-made objects as well as non-rigid or organic shape surfaces, and we believe it has potential for generative 3D modeling and predictive 3D tracking tasks.

{\small
\bibliographystyle{ieee}
\bibliography{egbib}
}

\end{document}